\documentclass[runningheads]{llncs}
\usepackage{eccv}
\usepackage{eccvabbrv}
\usepackage{bbding}

\usepackage{graphicx}
\usepackage{amsmath}
\usepackage{amssymb}
\usepackage{booktabs}
\usepackage{makecell}
\usepackage[export]{adjustbox}
\usepackage[normalem]{ulem}
\usepackage{textpos}  %
\usepackage{xcolor,colortbl}
\usepackage{pifont}

\usepackage{algorithmic}
\usepackage{wrapfig}
\usepackage[linesnumbered,ruled,vlined]{algorithm2e}

\SetKwComment{Comment}{\color{green!50!black}\# }{}

\newcommand{\var}{\texttt}

\SetKwProg{Function}{def}{:}{}

\SetKwProg{For}{for}{:}{}
\SetKwProg{If}{if}{:}{}

\usepackage{tabulary,multirow,overpic,xcolor,subfloat}
\usepackage[pagebackref=false, breaklinks=true, letterpaper=true, colorlinks,
            citecolor=citecolor, linkcolor=linkcolor, bookmarks=false]{hyperref}
\definecolor{citecolor}{HTML}{0071BC}
\definecolor{linkcolor}{HTML}{ED1C24}
\definecolor{lightblue}{rgb}{0.95, 0.95, 1}
\newcommand{\colorrect}[1]{\textcolor{#1}{\ding{110}}}

\newcommand{\gray}[1]{\textcolor{gray}{#1}}
\newcommand{\app}{\raise.17ex\hbox{$\scriptstyle\sim$}}

\newcolumntype{x}[1]{>{\centering\arraybackslash}p{#1pt}}
\newcolumntype{y}[1]{>{\raggedright\arraybackslash}p{#1pt}}

\newlength\savewidth\newcommand\shline{\noalign{\global\savewidth\arrayrulewidth
  \global\arrayrulewidth 1pt}\hline\noalign{\global\arrayrulewidth\savewidth}}
\newcommand{\tablestyle}[2]{\setlength{\tabcolsep}{#1}\renewcommand{\arraystretch}{#2}\centering\footnotesize}
\makeatletter\renewcommand\paragraph{\@startsection{paragraph}{4}{\z@}
  {.5em \@plus1ex \@minus.2ex}{-.5em}{\normalfont\normalsize\bfseries}}\makeatother

\newcommand\blfootnote[1]{\begingroup\renewcommand\thefootnote{}\footnote{#1}\addtocounter{footnote}{-1}\endgroup}

\DeclareMathAlphabet\mathbfcal{OMS}{cmsy}{b}{n}

\definecolor{Gray}{gray}{0.5}

\newcommand{\na}{\textcolor{gray}{ {n/a}}}

\newcommand{\modelname}{\textsc{SegIC}\xspace}

\newcommand{\sota}[0]{state-of-the-art\xspace}

\usepackage[capitalize]{cleveref}
\crefname{section}{Sec.}{Secs.}
\Crefname{section}{Section}{Sections}
\Crefname{table}{Table}{Tables}
\crefname{table}{Tab.}{Tabs.}

\definecolor{Gray}{gray}{0.93}

\crefname{section}{\S}{\S\S}
\crefname{subsection}{\S}{\S\S}
\crefformat{table}{Table~#2#1#3}
\crefformat{figure}{Figure~#2#1#3}
\crefformat{equation}{Equation~(#2#1#3)}
\crefformat{algorithm}{Algorithm~#2#1#3}
\crefformat{appendix}{Appendix~#2#1#3}

\makeatletter
\newcommand{\removelatexerrortwo}{\let\@latex@error\@gobble}
\makeatother
\newcommand{\myalgorithm}{%
\begingroup
\removelatexerrortwo
\begin{algorithm*}[H]
      \caption{Pseudo code for \modelname Mask Decoding.}
      \label{alg:mask_decoding}
      \scriptsize
          \Comment{Inputs:~Image Embedding $\mathbf{f}$; In-context Instructions $\mathbf{c} = \{\mathbf{a},\mathbf{p},\mathbf{v},\mathbf{m} \}$
          }
          \Comment{Variables:~Learnable Object Queries $\mathbf{q}$}
          \Comment{Functions:~Conv4ImgFeature(), Conv4ProgatedLabel(); 
          Proj4Pos(), Proj4Vis(), Porj4Meta();
          QuerySelfAttn(), Query2ImgAttn(), Image2QueryAttn(), output()}
  
  \Function{InContext\_Enhancement($\mathbf{f},\mathbf{a},\mathbf{p},\mathbf{v},\mathbf{m}$)}{
        \Comment{Project image feature and in-context propagated mask into the hidden space for mask decoding.}
        
        \var{$\mathbf{f}', \mathbf{a}'$=Conv4ImgFeature($\mathbf{f}$), Conv4ProgatedLabel($\mathbf{a})$};
        
        \Comment{Enhance image feature with in-context propagated mask.}
        
        \var{$\mathbf{f}'$=$\mathbf{f}'$+$\mathbf{a}'$};

        \Comment{Project other in-context instructions into the hidden space for mask decoding.}
        
        \var{$\mathbf{q^p}, \mathbf{q^v}, \mathbf{q^m}$=Proj4Pos($\mathbf{p}$),Proj4Vis($\mathbf{v}$),Proj4Meta($\mathbf{m}$)}
        
        \Comment{Enhance the object query by contacting with the hidden features of in-context instructions.}
        
        \var{$\mathbf{q}'$=Concat($\mathbf{q}, \mathbf{q^p}, \mathbf{q^v}, \mathbf{q^m}$)}

  }
  \Function{Mask\_Decoder($F$, $Q$)}{
    \var{$Q'$ = QuerySelfAttn($Q$)} \Comment{Query self-attention}

    \var{$Q^o$ = Img2QueryAttn($Q'$, $F$)} \Comment{Image-to-query cross-attention}
    
    \var{$F^o$ = Query2ImgAttn($F$, $Q_o$)} \Comment{Query-to-image cross-attention}

    \var{$O$ = output($F^o$, $Q^o$[0])} \Comment{Compute mask }
    
  }
  
  \Function{forward($\mathbf{f},\mathbf{a},\mathbf{p},\mathbf{v},\mathbf{m}$))}{
     \var{$\mathbf{f}'$, $\mathbf{q}'$ = InContext\_Enhancement($\mathbf{f},\mathbf{a},\mathbf{p},\mathbf{v},\mathbf{m}$)} \Comment{Enhance image feature and object query with in-context instructions.}
     
     \var{$Q^o$,$F^o$ = $\mathbf{q}'$, $\mathbf{f}'$} \Comment{Initialize variables for mask decoding}
     
     \For{$i$ in range(max\_iter)}{
    \var{$O$, $Q^o$,$F^o$ = Mask\_Decoder($Q^o$,$F^o$)}
     }
  }
\end{algorithm*}
\endgroup}

\makeatletter
\newcommand{\removelatexerror}{\let\@latex@error\@gobble}
\makeatother
\newcommand{\myalgorithmtwo}{
\begingroup
\removelatexerror
\begin{algorithm*}[H]
      {\scriptsize\caption{ Pseudo code for training pipeline.}}
      \label{alg:training_pipeline}
      \scriptsize
          \Comment{training set: mixed dataset $\mathbf{D}=\mathbf{D_{inst}} \cup \mathbf{D_{sem}}$}
  \Function{ICL\_Preprocess($\mathbf{data}$)}{
        \var{$\mathbf{I^t}$, $\mathbf{y^t}$, category} = $\mathbf{data}$

        \If{task\_type(data) == 'semantic'}{
            
            $\mathbf{I^r}$, $\mathbf{y^r}$ = CategoryAwareSample(\var{$\mathbf{D_{sem}}$, category}) 
            
            $\mathbf{meta}$ = \var{'a photo of a \{category\}.' }  
            }
        \If{task\_type(data) == 'instance'}{
            $\mathbf{I^r}$, $\mathbf{y^r}$ = DataAug($\mathbf{I}^t$, $\mathbf{y}^t$) \Comment{Individual data aug to build a different view as reference} 
            
            $\mathbf{meta}$ = \var{'please segment the instances.'} 
            }
  }
  \vspace{-1mm}
  \Function{train\_epoch(model, $\mathbf{D}$)}{
    \For{data in $\mathbf{D}$}{
    
    \var{$\mathbf{I^t}$, $\mathbf{y^t}$, $\mathbf{I^r}$, $\mathbf{y^r}$, $\mathbf{meta}$ = ICL\_Preprocess(data)}
    
    \var{loss} = model($\mathbf{I^t}$, $\mathbf{y^t}$, $\mathbf{I^r}$, $\mathbf{y^r}$, $\mathbf{meta}$) 
    }
  }
\vspace{-1.8mm}
\end{algorithm*}
\endgroup}

\begin{document}

\title{\Large 
\modelname: Unleashing the Emergent Correspondence for In-Context Segmentation
}
\titlerunning{Unleashing the Emergent Correspondence for In-Context Segmentation}

\author{
  Lingchen Meng$^{1,2}$~\hspace{10pt}
  Shiyi Lan$^{3}$~\hspace{10pt}
  Hengduo Li$^{4}$~\hspace{10pt} \\
  Jose M. Alvarez$^{3}$~\hspace{10pt}
  Zuxuan Wu$^{1,2\dagger}$~\hspace{10pt} 
  Yu-Gang Jiang$^{1,2}$
}
\authorrunning{L. Meng et al.}

\institute{
$^{1}$Shanghai Key Lab of Intell. Info. Processing, School of CS, Fudan University \\
$^{2}$Shanghai Collaborative Innovation Center of Intelligent Visual Computing \\
$^{3}$NVIDIA~\hspace{10pt} 
$^{4}$University of Maryland
}

\makeatletter
\let\@oldmaketitle\@maketitle%
\renewcommand{\@maketitle}{\@oldmaketitle%
\centering
    \includegraphics[width=\linewidth]{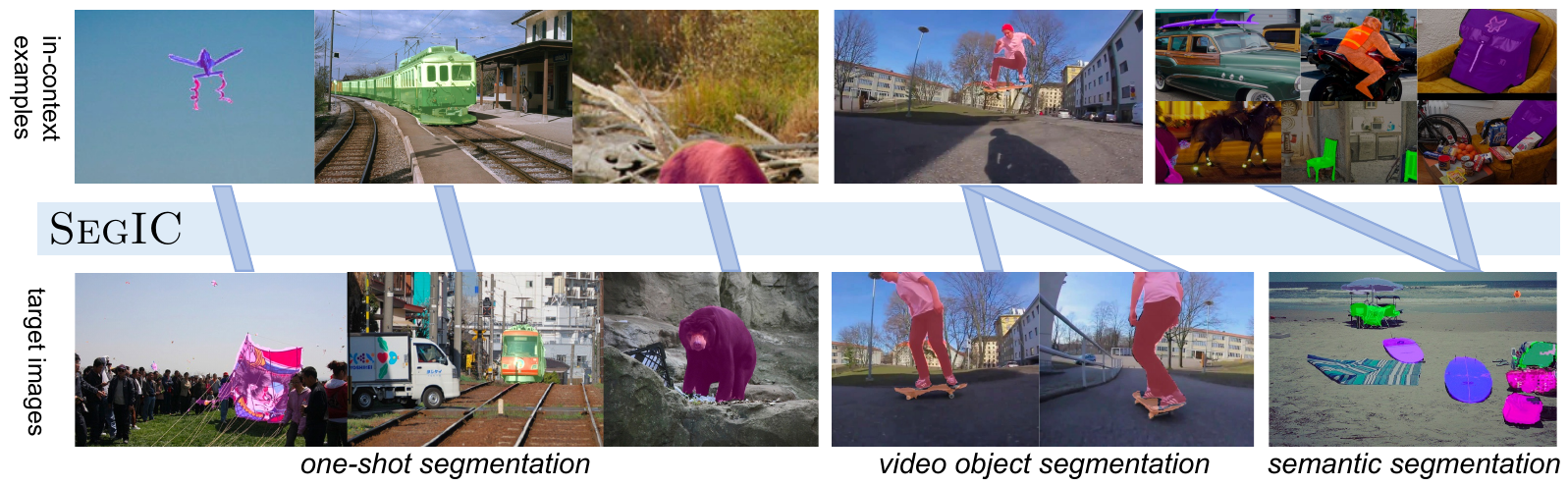}
    \captionof{figure}{%
    \textbf{Qualitative results of \modelname.} 
    {
    \modelname segments target images (the bottom row) according to a few labeled example images (top row, linked by \includegraphics[height=2ex,valign=B]{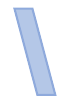} in the figure), termed as ``in-context segmentation''. \modelname unifies various segmentation tasks via different types of in-context samples, including those annotated with one mask per sample (one-shot segmentation),  annotated with a few masks per sample (video object segmentation), and the combination of annotated samples (semantic segmentation)}
    }
    \vspace{-10mm}
    \label{fig:teaser}
\bigskip}
\makeatother

\maketitle

\blfootnote{$^{\dagger}$ Corresponding author.}

\begin{abstract}

In-context segmentation aims at segmenting novel images using a few labeled example images, termed as ``in-context examples'', exploring content similarities between examples and the target.
The resulting models can be generalized seamlessly to novel segmentation tasks, significantly reducing the labeling and training costs compared with conventional pipelines.
However, in-context segmentation is more challenging than classic ones requiring the model to learn segmentation rules conditioned on a few samples.
Unlike previous work with ad-hoc or non-end-to-end designs, we propose \modelname, an end-to-end \textbf{seg}ment-\textbf{i}n-\textbf{c}ontext framework built upon a single vision foundation model (VFM). In particular, \modelname leverages the emergent correspondence within VFM to capture dense relationships between target images and in-context samples. As such, information from in-context samples is then extracted into three types of instructions, \ie geometric, visual, and meta instructions, serving as explicit conditions for the final mask prediction. 
\modelname is a straightforward yet effective approach that yields state-of-the-art performance on one-shot segmentation benchmarks.
Notably, \modelname can be easily generalized to diverse tasks, including video object segmentation and open-vocabulary segmentation. 
Code will be available at \url{https://github.com/MengLcool/SEGIC}.
\end{abstract}

\vspace{-2mm}
\section{Introduction}

Modern advancements in deep learning have established a routine process for addressing visual perception challenges, typically involving data collection, model training, and deployment. While this pipeline is highly effective, it invariably demands additional effort in data acquisition and model tuning when adapting to new domains. Although researchers have been seeking to learn generic representations with pre-training, the resulting models have to be fine-tuned on the target domain for improved performance.

In contrast, the success of large language models (LLMs)~\cite{gpt1, gpt2, gpt3, instruct-gpt} in Natural Language Processing (NLP) offers an alternative approach. These models are trained on vast datasets, handling various NLP tasks through next-token prediction guided by prompts~\cite{instruct-gpt}. A key strength of LLMs is their ability to learn from a few examples, a process known as in-context learning (ICL). This enables them to adapt to various tasks with a small and varied set of instructions without requiring extensive fine-tuning or retraining~\cite{gpt3,instruct-gpt}. The success of ICL in NLP highlights the potential for applying similar strategies in visual perception tasks.

While appealing, ICL in vision is particularly challenging as vision tasks are significantly different regarding inputs (2D/3D), outputs (one-hot labels/bounding boxes), and specialized architectures. Recent advances in vision generalist models~\cite{x-decoder, seem, uniperceiver-v2} suggest different levels of segmentation tasks, \ie instance, semantic, and video, can be unified within the same output space. This motivates us to explore ICL using segmentation as a testbed and investigate whether current vision models can be easily generalized. While there are a few previous attempts
on ICL for segmentation, they either have fallen short in performance due to implicit modeling~\cite{painter, seggpt} or have employed heavy and non-end-to-end pipelines~\cite{matcher, persam, dift}, which are less effective and efficient.

At the heart of ICL for NLP tasks is mining the relationships among different words and then propagating labels from a few task-specific question-answer pairs, namely in-context samples to the target one~\cite{xie2021explanation,kossen2023context,balavzevic2023towards}. We argue that in vision tasks, the similar entity that facilitates label propagation from in-context samples to novel samples is establishing dense correspondences between images.
Although dense visual correspondences are difficult to obtain before the era of foundation models, recent studies~\cite{dift,dino} have shown that high-quality correspondence emerges in visual foundation models (VFMs)~\cite{dino, dinov2, clip,sd}.

In light of this, we introduce \modelname, an end-to-end \textbf{seg}ment-\textbf{i}n-\textbf{c}ontext framework without the need for sophisticated handcrafted prompt design. Specifically, our framework is built upon a single frozen vision foundation model followed by a lightweight mask decoder. We leverage the emergent correspondence of the VFM to establish dense correspondences across target images and in-context samples. Based on that, we extract in-context information into three types of instructions: geometric, visual, and meta instructions. By explicitly utilizing these instructions, our model demonstrates remarkable generalization capabilities with low training costs across diverse segmentation tasks, as evidenced by extensive qualitative and quantitative experiments. We summarize our contributions in threefold:
\begin{itemize}
    \item  We introduce \modelname, a simple yet effective in-context segmentation framework, exploring the strong emergent correspondence encoded in VFMs.
    \item We design geometric, visual, and meta instructions that explicitly transfer knowledge from in-context samples to the target to facilitate in-context segmentation without tuning the parameters of vision foundation models. 
    \item \modelname demonstrates \sota performance on COCO-20$^i$, FSS-1000 and recent LVIS-92$^i$. Moreover, we conduct a comprehensive study on vision foundation models of various pre-text tasks, model sizes, and pre-training data.
    Furthermore, \modelname achieves competitive performance on novel tasks including video object segmentation and open-vocabulary segmentation, without ever seeing their training data. 
\end{itemize}

\section{Related Work}
\noindent\textbf{Vision foundation models.} 
Recent years have witnessed great progress in large-scale vision pre-training~\cite{clip, simclr, moco, dino, beit, mae, sam}, serving as the cornerstone for high-capacity foundation models. These pre-training approaches can be broadly categorized into two directions: vision-only  pretraining~\cite{mae, moco, simclr, dino, dinov2, beit} and vision-language pre-training~\cite{clip, align, albef, coca, sd}.
For vision-only pre-training, models aim to distinguish image/patch-level entities from different views~\cite{dino, simclr, moco} or reconstruct masked images~\cite{beit, moco} from raw images. In vision-language pre-training, models strive to align cross-modal features into a unified visual-semantic space~\cite{clip, align, coca, albef}, showcasing great open-set performance due to the transferable capabilities of language. Unlike these approaches that perform pre-training in an unsupervised or weakly supervised manner, SAM~\cite{sam} is pre-trained on a huge amount of labeled segmentation data with precise prompts about locations. In this paper, we conduct extensive experiments using three types of pre-trained models as backbones to explore their potential for in-context segmentation. Interestingly, we observe that models with higher zero-shot semantic and geometric correspondence performance are more likely to be effectively utilized in our \modelname framework for in-context segmentation.

\vspace{0.05in}
\noindent\textbf{Unified vision downstream models.}
Unified vision downstream models have recently drawn significant attention due to their generalization capabilities and flexibilities. Unlike previous specialized vision models designed for specific datasets \cite{faster_rcnn, fcn, maskrcnn, deeplab, upernet}, vision generalists are tailored to handle multiple datasets~\cite{unidet, dhub, dataseg, richsem} and a wide range of tasks~\cite{x-decoder, seem, oneformer} within a single yet unified model. Recently, many studies~\cite{x-decoder, seem, oneformer, sam, painter, seggpt} have focused on developing techniques that unify segmentation tasks. 
In a similar spirit, our work also builds upon a unified output space for segmentation tasks. However, our goal is different---we aim to perform in-context segmentation that allows a model to effectively segment novel
entities conditioned on a few samples.

\vspace{0.05in}
\noindent\textbf{Visual correspondence.}
Establishing visual correspondences between different images is vital for various computer vision tasks. Traditionally, computing correspondences is based on hand-designed features like SIFT~\cite{sift} and SURF~\cite{surf}.  Deep models~\cite{lee2021patchmatch,kim2022transformatcher, wang2019learning} can also learn correspondence in a supervised manner. 
More recently, it has been shown features from large foundation models encode dense visual correspondence clues~\cite{moco,mae,dino,dinov2,beit}. In this study, we discover a profound connection between correspondence and in-context segmentation---correspondence acts as explicit guidance, linking the target image with in-context images, thus facilitating label propagation for in-context segmentation.

\vspace{0.05in}
\noindent\textbf{In-context learning.}
For the first time, GPT-3~\cite{gpt3} introduces a new learning paradigm known as in-context learning, which unifies various NLP tasks as text completion or question-answering tasks using provided prompts and examples. This approach enables language models to handle various tasks, including novel ones, by leveraging task examples, without requiring re-training or fine-tuning.  Recent studies~\cite{painter, seggpt, matcher, bar2022visual, bai2023sequential} explore this mechanism in vision tasks. Painter~\cite{painter} and SegGPT~\cite{seggpt} aim to achieve in-context segmentation through in-painting. They build upon the Mask Image Modeling (MIM) framework~\cite{mae} to concatenate images and predictions into a 2$\times$2 mosaic and make predictions by recovering the masked areas. In their pipelines, the vision backbone serves as both an image encoder and a mask decoder, which incurs significant computational costs. Moreover, they struggle to effectively leverage pre-trained models due to input shifts, leading to increased convergence challenges. Other approaches~\cite{persam,matcher} attempt using in-context segmentation via prompting SAM~\cite{sam}. They build upon cross-image correspondences between in-context examples and target images by additional pre-trained models to generate prompts for SAM. However, these methods employ a two-stage pipeline, introducing redundancy and repeated computations. Consequently, if the model encounters limitations in one stage, it negatively impacts the final performance. In this work, we build an end-to-end in-context segmentation framework, leveraging the emergent correspondence using a single vision foundation model.

\section{Approach}

\begin{figure*}[!t]
    \centering
    \includegraphics[width=\linewidth]{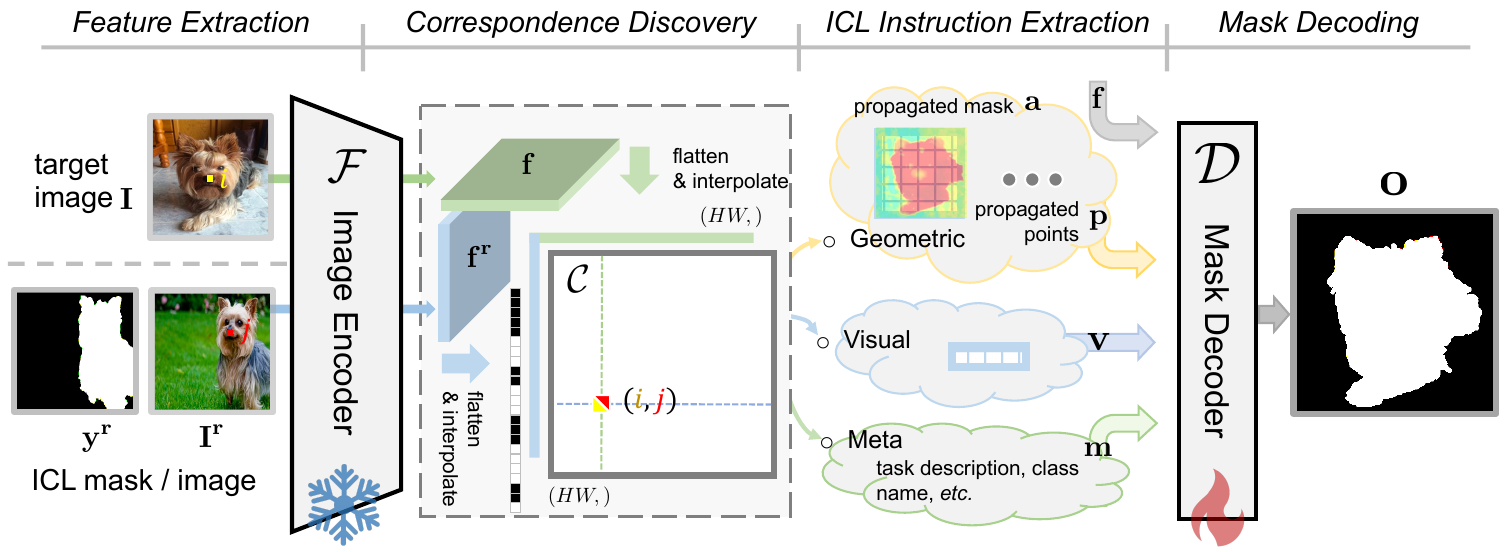}
    \caption{\textbf{Architecture overview.} \modelname is built upon a \textbf{frozen} vision foundation model, consisting of four stages: (1) feature extraction; (2) correspondence discovery (Section~\cref{sec:dense_corr}); (3) in-context instruction extraction (Section~\cref{sec:icl_features}); (4) mask decoding (Section~\cref{sec:mask_decoding}).}
    \label{fig:arch}
    \vspace{-5mm}
\end{figure*}

In-context learning equips a model with the ability to learn from example images, namely ``in-context examples'', as humans, which has demonstrated great potential in NLP tasks~\cite{gpt3,instruct-gpt}.  This process is akin to how humans intuitively grasp and replicate complex patterns from just a few guides, rapidly generalizing to new examples. In this paper, our goal is to establish an end-to-end in-context segmentation framework, enabling a model to effectively segment novel entities beyond the training set, such as video objects, with low training costs.

Formally, given a target image $\mathbf{I} \in \mathbb{R}^{3\times H\times W}$ where $H$ and $W$ represent the height and width, the goal is to segment a binary mask $\mathbf{y}\in \mathbb{R}^{\{0,1\}{H\times W}}$ conditioned on $K$ in-context examples $ \{ \mathbf{x^r} \}^K = \{( \mathbf{I^r}, \mathbf{y^r})\}^K$, where $\mathbf{I^r}, \mathbf{y^r}$ are the image and ground-truth mask of an in-context example. 

Our approach consists of four stages: feature extraction, correspondence discovery, in-context instruction extraction, and mask decoding, as shown in \cref{fig:arch}.
The feature extraction stage follows the common practice in visual perception tasks where we use a pre-trained visual foundation model to obtain the feature maps of both in-context and target images. Subsequently, we compute the dense correspondences based on their semantics and geometry (Section~\cref{sec:dense_corr}), providing explicit guidance of the segmentation rules.
Then, we extract in-context instructions based on the in-context samples and the dense correspondences (Section~\cref{sec:icl_features}).
Finally, \modelname uses a lightweight in-context mask decoder $\mathcal{D}$ to produce segmentation masks for the desired target, leveraging the extracted information from in-context examples (Section~\cref{sec:mask_decoding}).

\subsection{Dense Correspondence Discovery}
\label{sec:dense_corr}

To establish the relations between the target image $\mathbf{I}$ and the
 reference image $\mathbf{I^r}$ (considering one reference example for brevity), we extract dense cross-image correspondences at the pixel level. For this purpose, we leverage pre-trained VFMs due to their powerful generalization capability and emergent correspondence properties.
Specifically, we first extract the visual features of $\mathbf{I}$ and $\mathbf{I^r}$ with a vision foundation model, then apply the cosine distance function to compute the patch-level distance map between the two images as shown in~\cref{fig:arch}. We further obtain the pixel-level correspondences by interpolating the patch-level distance map to the size of the original image as follows:
\begin{equation} \label{eq:corr}
\begin{split}
    \mathbf{f}, \hspace{1mm}\mathbf{f^r} &= \mathcal{F}(\mathbf{I}), \hspace{1mm}\mathcal{F}(\mathbf{I^r}) \\
    \mathcal{C} &= \texttt{Upsample}\left(\texttt{Dist}(\mathbf{f}, \mathbf{f^r})\right),
\end{split}
\end{equation}
where $\mathcal{F}$ indicates the vision foundation model we use and $\mathbf{f}, \mathbf{f^r} \in \mathbb{R}^{c\times h\times w}$ are the extracted features of the target and the reference images, respectively; $c, h, w$ indicate the dimensions of the feature maps. $\texttt{Dist}$ denotes the distance function (for which we use cosine distance) and $\texttt{Upsample}$ is the interpolation function. 
$\mathcal{C}\in \mathbb{R}^{HW\times HW}$ denotes the calculated dense correspondences between the target image and the in-context image.

\subsection{In-context Instruction Extraction}
\label{sec:icl_features}
After obtaining dense correspondences, the question becomes how to utilize the in-context information, including in-context examples and dense correspondences, as instructions to guide the segmentation process. Here we extract in-context instructions based on ideas from NLP tasks~\cite{instruct-gpt, kossen2023context}.
Ideally, the representation for in-context information should clearly articulate how segmentation should be executed on the target image while being concise and efficient for effective segmentation. 
To this end, we decouple and encode the in-context information into three individual in-context instructions as shown in~\cref{fig:arch}: 1) geometric instructions; 2) visual instructions; and 3) meta instructions, each of which will be elaborated below:

\vspace{0.05in}
\noindent\textbf{Geometric instructions} aim to provide a coarse location estimation of the target mask. 
Although we only have the mask annotation for the reference image, we can propagate the label from the reference to the target to obtain a propagated mask by exploring the dense correspondences $\mathcal{C}$:
\begin{equation} \label{eq:feat_geo}
\begin{split}
\mathbf{a} &= \mathbf{y^r}/ \|\mathbf{y^r}\| \cdot\mathcal{C}  \in \mathbb{R}^{H\times W}  \\
\mathbf{p} &= \texttt{PE}(\texttt{Topk}(\mathbf{a})),
\end{split}
\end{equation}
where $\mathbf{a}$ represents the propagated coarse mask for the target image.
As shown in~\cref{eq:feat_geo}, by performing matrix multiplication, we gather and average the dense correspondences based on the positive points in the reference mask. This process is analogous to propagating labels from the reference image to the target image through dense correspondences, making $\mathbf{a}$ a form of dense geometric instruction. Additionally, based on $\mathbf{a}$, we employ $\mathrm{Topk}$ (as referred to in~\cref{eq:feat_geo}) to select the top-k points with the highest values in $\mathbf{a}$, indicating the locations most likely relevant to the target mask. We further encode the 2D coordinates into high-dimensional vectors using cosine positional encoding $\mathrm{PE}$ as in~\cite{transformer, tancik2020fourier}, resulting in a type of sparse geometric instruction. Overall, $\mathbf{a}$ and $\mathbf{p}$ together provide geometric information extracted from in-context examples.

\vspace{0.05in}
\noindent\textbf{Visual instructions} indicate the visual clues of the target entity.
We utilize mask pooling~\cite{odise} on reference image features to extract visual salient clues, $\mathbf{v}$:
\begin{equation} \label{eq:feat_vis}
\begin{split}
\mathbf{v} &= \mathbf{y^r}/ \|\mathbf{y^r}\| \cdot\mathbf{f^r}. \\
\end{split}
\end{equation}
By doing so, only information relevant to the reference mask, including low-level (texture, appearance) and high-level (semantic) clues, is used.

\vspace{0.05in}
\noindent\textbf{Meta instructions} indicate other clues implicitly provided by in-context examples, such as task descriptions, class names, \etc. We uniformly treat them as languages and encode them with a pre-trained language model, following~\cite{oneformer, grounding-dino, sam}.
\begin{equation} \label{eq:sem}
\begin{split}
\mathbf{m} &= \mathcal{F}^t(\mathbf{meta})  \\
\end{split}
\end{equation}
where $\mathcal{F}^t$ is the pre-trained CLIP text encoder~\cite{clip}, $\mathbf{meta}$ indicates the meta information and $\mathbf{m}$ is the meta feature.

Finally, we use $\mathbf{c} = \{ \mathbf{a},\mathbf{p},\mathbf{v},\mathbf{m} \}$ to denote the set of in-context instructions derived from reference samples, which can be further used for producing the segmentation mask, as will be introduced in section~\cref{sec:mask_decoding}.

\subsection{Mask Decoding}
\label{sec:mask_decoding}
In this section, we discuss how to predict the segmentation masks in target images following the aforementioned in-context instructions. In particular, we use a query-based mask decoder $\mathcal{D}$ due to its high performance in segmentation tasks and flexibility~\cite{detr,maskformer,mask2former}. Formally, the decoder network takes the target image feature $\mathbf{f}$ and in-context instructions $\mathbf{c}$ as input, as well as a learnable query $\mathbf{q}$, and outputs a mask $\mathbf{o}$ as follows:
\begin{equation} \label{eq:icl_decoding}
\begin{split}
    \mathbf{o} = \mathcal{D}(\mathbf{f}, \mathbf{c} ;\hspace{1mm} \mathbf{q}). \\
\end{split}
\end{equation}
Unlike previous designs that use a set of object queries~\cite{detr, maskformer, mask2former} for prediction, we only initialize one query since we just need to predict one mask in-context conditioned on the in-context instructions as shown in~\cref{fig:arch}.

To prepare these instructions for mask decoding, we first project them into the latent space used by the decoder.
We categorize the in-context instructions into two types according to their spatial property: instructions with spatial shapes (\ie the propagated coarse mask $\mathbf{a}$) and without spatial shapes (\ie $\mathbf{p}, \mathbf{v}, \mathbf{m}$).
For the spatial instructions, we employ a series of convolutional layers $\mathcal{M}$ to encode it into the image feature space; for the non-spatial instructions, we use projection layers $\mathcal{P}$ to project them into the query feature space:
\begin{equation} \label{eq:cond_proj}
\begin{split}
\mathbf{q^{p}},\mathbf{q^{v}},\mathbf{q^{m}} &= \mathcal{P}^{p}(\mathbf{p}),\hspace{1mm}
\mathcal{P}^{v}(\mathbf{v}),\hspace{1mm} \mathcal{P}^{m}(\mathbf{m}) \\
\mathbf{a'} &= \mathcal{M}(\mathbf{a}) 
\end{split}
\end{equation}
where $\mathcal{P}^{p},\mathcal{P}^{v},\mathcal{P}^{m}$ indicate the projection layers for $\mathbf{p},\mathbf{v},\mathbf{m}$, respectively. $\mathbf{a'}$, $\mathbf{q^{p}}$, $\mathbf{q^{v}}$ and $\mathbf{q^{m}}$ are the projected in-context instructions.

Furthermore, we inject the projected in-context instructions into the decoding stage to guide the decoder to segment in-context. 
Similarly, for the spatial features, we add them to image features, such that they are aware of the coarse mask produced by reference samples.
For the non-spatial features, we concatenate the initial query with them, which allows a deeper interaction via a self-attention mechanism in the decoder:
\begin{equation} \label{eq:icl_seg}
\begin{split}
    \mathbf{f'} &=\mathbf{f} + \mathbf{a'} \\
    \mathbf{q'} &= \mathrm{Concat}(\mathbf{q}, \mathbf{q^{l}},\mathbf{q^{s}},\mathbf{q^{m}}) \\
    \mathbf{o} &= \mathcal{D}(\mathbf{f'}; \mathbf{q'}) \\
\end{split}
\end{equation}
Finally, the mask prediction $\mathbf{o}$ is produced based on image features~$\mathbf{f'}$ conditioned on instructions, and query features~$\mathbf{q'}$ as shown in~\cref{eq:icl_seg}. For more details, please refer to our supplementary.

\subsection{Training Pipeline}
During training, we freeze all the parameters of the VFM and only leave the newly introduced mask decoder trainable. 
We employ a linear combination of a dice loss~\cite{dice} and a binary cross-entropy loss for our mask loss: 
$L_{mask}=\lambda_{ce}L_{ce}+\lambda_{dice}L_{dice}$. It is worth noting that we calculate the segmentation loss on $K$ selected points using importance sampling following~\cite{mask2former,pointrend} instead of the whole image to save memory cost.
To further improve the robustness toward noisy in-context samples, we introduce two strategies into our training recipe, namely ``context reversion'' and ``negative entity augmentation''.

\vspace{0.05in}
\noindent\textbf{Context reversion.}
We artificially introduce noisy context during training to improve the robustness. To simulate situations where in-context examples are inaccurate, we propose ``context reversion'': swapping the target and reference images---we use the prediction of the target image as an in-context example. The noisy context introduces randomness during training and hence can improve robustness.

\vspace{0.05in}
\noindent\textbf{Negative entity augmentation.}
In tasks like video object segmentation~(VOS), in-context examples for different entities in the same image are mutually exclusive. Taking the case of video object segmentation in~\cref{fig:teaser} as an example, the person and the skateboard are exclusive in one image.
Thus, entities that are not of interest can serve as negative samples, indicating that they are not relevant to the target. We augment the in-context instructions with these negative entities for a better result.

\section{Experiments}

\subsection{Training Data}
We train \modelname on semantic and instance segmentation datasets. Unlike previous sophisticated task-specific designs for multiple datasets/tasks, our method offers a simple approach that distinguishes tasks by in-context examples.

\vspace{0.05in}
\noindent\textbf{COCO}~\cite{coco} is a widely used instance/semantic segmentation with 83K training samples and 41K samples of 80 categories. We use the 2014 split version to be consistent with the COCO-20$^i$ one-shot semantic segmentation benchmark.

\vspace{0.05in}
\noindent\textbf{ADE20k}~\cite{ade20k} is a widely used semantic segmentation dataset for 150 categories, with 20K training images.

\vspace{0.05in}
\noindent\textbf{LVIS}~\cite{lvis} is a large instance segmentation dataset containing 1000+ categories with $\sim$100K training images. 

\subsection{Training Details}\label{sec:detail}
We implement \modelname in PyTorch and use 8 V100 GPUs for most of our
experiments. We use a batch size of 32 (4 per GPU) in total, 1 point for point instruction, and 12544 points per mask to calculate mask loss following~\cite{mask2former}. We merge the COCO instances for its semantic segmentation. Please refer to our supplementary material for the detailed training pipeline.
Thanks to the frozen backbone, our method is extremely memory-efficient (with less than 10G memory cost in most of our experiments).
\modelname is a single unified model that is jointly trained on mixed datasets, while evaluated on various datasets, separately. We utilize DINOv2~\cite{dinov2} as the default vision foundation model, with a ViT-B for all ablations, and ViT-L/G for the main experiments. We employ an AdamW~\cite{adamw} optimizer with a weight decay of $1\mathrm{e}{-4}.$ We set an initial learning rate as $1\mathrm{e}{-4}$ and multiply 0.1 at the 10 epoch during training. Each dataset is sampled uniformly with 160K samples per epoch in the main experiments and 80K samples for the ablations. 
We perform data augmentations on target images and reference images respectively. We use large-scale jittering augmentation for semantic segmentation datasets, and normal data augmentations, including random resizing cropping, color jittering, and random horizontal flipping, for instance segmentation datasets. We random sample 1 mask per image for semantic segmentation datasets during training, while up to 10 masks for instance segmentation. The size of a single image is cropped/padded to 896$\times$896.

\begin{table}[!h]
  \centering
  \tablestyle{1.5pt}{1.2}
  \scriptsize
    \caption{ \textbf{Main results on several segmentation benchmarks.} Param$^t$ indicates the number of trainable parameters. $^{\dag}$ indicates relying SAM; * indicates that needs additional fine-tuning, \# indicates that classes within images are known during inference; \na\ indicates the model does not have capability for the task and - indicates that do not have reported number. For FSS-1000 and VOS datasets, we highlight the performance trained on corresponding datasets in \textcolor{gray}{gray} and zero-shot results in black.
  }
  \vspace{-3mm}
  \resizebox{\linewidth}{!}{
  \begin{tabular}{lrc| ccc | cc | cc | cc}
  \multirow{3}{*}{Method} &  \multirow{3}{*}{Param$^{l}$} && \multicolumn{3}{c}{\textit{one-shot segmentation}} & \multicolumn{2}{c}{\textit{video object segmentation}}  & \multicolumn{2}{c}{\textit{semantic seg}} & \multicolumn{2}{c}{\textit{open-vocab seg}}\\
   &&& COCO-20$^i$ & FSS-1000 & LVIS-92$^{i}$ & DAVIS-17 & YVOS-18 & COCO & ADE20k  & PC-459 & A-847\\
   &&& mean mIoU & mIoU & mean mIoU & $\mathcal{J}$\&$\mathcal{F}$ & G & mIoU & mIoU & mIoU& mIoU\\
  \shline
  \multicolumn{2}{l}{\emph{\textcolor{darkgray}{few-shot seg specialist}}} &&&&&&&&& \\
  HSNet (RN50)~\cite{hsnet} & 28M && 41.7 & \gray{86.5} & 17.4& \na & \na & \na & \na & \na & \na \\
  VAT (RN50)~\cite{VAT} & 52M &&  42.9 & \gray{90.3} & 18.5 & \na & \na & \na & \na & \na & \na \\
  FPTrans (B)~\cite{fptrans} & 101M &&  56.5 & - & - & \na & \na & \na & \na & \na & \na \\
  \hline
  \multicolumn{2}{l}{\emph{\textcolor{darkgray}{VOS specialist}}}  &&&&&&&&& \\
  AGAME~(RN101)~\cite{agame} & - && \na & \na & \na & \textcolor{gray}{70.0} & \textcolor{gray}{66.0} & \na & \na & \na & \na  \\ 
  SWEM~(RN50)~\cite{swem} & 58M && \na & \na & \na & \textcolor{gray}{84.3} & \textcolor{gray}{82.8} & \na & \na & \na & \na  \\ 
  XMem~(RN50)~\cite{xmem} & 62M && \na & \na & \na & \textcolor{gray}{87.7} & \textcolor{gray}{86.1} & \na & \na & \na & \na  \\ 
  \hline 
  \multicolumn{2}{l}{\emph{\textcolor{darkgray}{semantic seg specialist}}}  &&&&&&&&& \\
  MaskFormer~(L)~\cite{maskformer} & 212M && \na & \na & \na & \na & \na &64.8&54.1 & \na & \na  \\ 
  Mask2Former~(L)~\cite{mask2former} & 216M && \na & \na & \na & \na & \na &67.4&56.1 & \na & \na  \\ 
  MaskDINO~(L)~\cite{maskdino} & 225M && \na & \na & \na & \na & \na & - & 56.6 & \na & \na  \\ 
  \hline
  \multicolumn{2}{l}{\emph{\textcolor{darkgray}{segmentation generalist}}} &&&&&&&&& \\
  OneFormer~(L)~\cite{oneformer} & 237M && \na & \na & \na & \na & \na &67.4&57.7&-&- \\ 
  UNINEXT~(L)~\cite{uninext} & 340M && \na & \na & \na &\textcolor{gray}{77.2}&\textcolor{gray}{78.1}& - & -&-&- \\
  X-decoder~(L)~\cite{x-decoder}  & 341M && \na & \na & \na & \na & \na & 67.5 & 58.1&29.6&9.2\\
  SEEM~(L)~\cite{seem} & 341M &&-&-&-& 58.9 & 50.0 & 67.6 & -&-&-\\
  \hline
  \multicolumn{2}{l}{\emph{\textcolor{darkgray}{in-context generalist}}} &&&&&&&&& \\
  Painter~(L)~\cite{painter}  & 354M && 33.1 & 61.7 & 10.5 & 34.6 & 24.1 &-&49.9&-&-\\
  SegGPT~(L)~\cite{seggpt} & 354M && 56.1 & 85.6 & 18.6 & 75.6 & 74.7 &-&*39.6&-&-\\
  PerSAM$^{\dag}$~(H)~\cite{persam} & 0 && 23.0 & 71.2 & 11.5 & 60.3 &-& \na & \na & \na & \na \\
  PerSAM-F$^{\dag}$~(H)~\cite{sam} & 2 && 23.5 & 75.6 & 12.3& 71.9 & -& \na & \na & \na & \na \\
  Matcher$^{\dag}$~(H+G)~\cite{matcher}  &0 && 52.7 & 87.0 & 33.0 & 79.5 & -& \na & \na & \na & \na \\
  \hline
  \modelname~(L) & 5M &&76.1&86.8& 44.6 & 71.4& 62.7 & \#72.9 & \#55.5&\#33.5&\#18.9 \\
  \modelname~(G) & 5M && 74.5&88.4& 47.8 &  73.7& 65.4& \#74.0 & \#59.0 & \#34.9 & \#20.1\\
  \end{tabular}
  }
\vspace{-9mm}
\label{tab:main_result}
\end{table}

\subsection{Main Results}
For the main experiments, we compare \modelname with other specialist/generalist methods on several benchmarks of different segmentation tasks. We use \modelname of DINOv2~\cite{dinov2} of large and giant versions as the backbone for the main experiments.

\vspace{0.05in}
\noindent\textbf{One-shot semantic segmentation.} 
To demonstrate the generalization capability from known categories to unknown ones, we evaluate \modelname on one-shot semantic segmentation benchmarks. 
Following SegGPT~\cite{seggpt}, we evaluate \modelname in two one-shot semantic segmentation settings: in-domain using COCO-20$^i$\cite{coco20i} (the training set of COCO-20$^i$ is a subset of ours) and out-of-domain with FSS-1000~\cite{fss} (without seeing any training samples of FSS-1000). As depicted in~\cref{tab:main_result}, \modelname has achieved state-of-the-art performance in both of these settings. To ensure a fair comparison, we report in-domain performance of specialist models for COCO-20$^i$ reported in~\cite{seggpt}.
Additionally, for FSS-1000, we highlight the performance trained on FSS-1000 in gray and zero-shot results in black. Notably, \modelname outperforms previous generalist models by a significant margin (more than 20 of mean mIoU) on COCO-20$^i$ and achieves competitive results that are very close to specialist models on FSS-1000, even without ever being trained on it. Furthermore, we conduct experiments on LVIS-92$^i$~\cite{matcher}, a more challenging one-shot benchmark built upon LVIS~\cite{lvis}. On LVIS-92$^i$, \modelname surpasses Matcher, the previous SoTA, by a large margin (from 33.0 to 47.8).

\begin{wraptable}{r}{6.5cm}
      \vspace{-10mm}
      \centering
      \tablestyle{4pt}{1.2}
      \scriptsize
    \caption{ \textbf{Comparisons on one-shot COCO-20$^i$.} \ddag\ indicates that is jointly trained on the COCO-\textbf{excluded} datasets.}
      \begin{tabular}{l | cccc|c}
      Method & F0 &  F1  & F2 & F3 & mean\\
      \shline
      HSNet~\cite{hsnet} & 37.2 & 44.1 & 42.4 & 41.3 & 41.2 \\
      VAT~\cite{VAT} & 39.0 & 43.8 & 42.6 & 39.7 & 41.3 \\
      FPTrans~\cite{fptrans} & 44.4 & 48.9 & 50.6 & 44.0 & 47.0\\
      MSANet~\cite{msanet} & 47.8 & 57.4 & 48.7 & 50.5 & 51.1\\
      \hline
      \modelname & 55.8 & 54.7 & 52.4 & 51.4 & 53.6\\
      \modelname \ddag & \textbf{62.3} & \textbf{62.5} & \textbf{63.3} & \textbf{60.9} & \textbf{62.3} \\
      \end{tabular}
    \vspace{-5mm}
    \label{tab:coco_20i}
\end{wraptable}
Since we include the evaluation categories in the training process for COCO-20$^i$ in~\cref{tab:main_result}, for a rigorous comparison, we follow its training setting that trains and tests our model on 4 splits~\cite{coco20i} separately, avoiding seeing categories for evaluation.
As shown in~\cref{tab:coco_20i}, our method still achieves \sota performance on COCO-20$^i$. 
Furthermore, as shown in the last row of~\cref{tab:coco_20i}, by joint training on COCO-excluded datasets (including ADE20k, LVIS, and FSS-1000), there is a significant performance gain across all splits. This further demonstrates the effectiveness of our in-context generalization capabilities.

Overall, our best model surpasses all previous segmentation generalist models across all one-shot segmentation benchmarks, demonstrating its effectiveness.

\vspace{0.05in}
\noindent\textbf{Zero-shot video object segmentation.} 
Video object segmentation (VOS) aims to segment specific objects in video frames. In this work, we focus on the semi-supervised VOS setting~\cite{davis,youtube}, where the masks that appeared first time are given as references. 
We evaluate \modelname without any fine-tuning on video datasets to demonstrate our generalization capabilities.
We choose two commonly used VOS datasets: DAVIS-17~\cite{davis} and YouTube-VOS-18~\cite{youtube}. We export two metrics commonly used in VOS for evaluation: the $\mathcal{J}\&\mathcal{F}$ score for DAVIS-17 and $G$ score for YouTube-VOS-18, with their official evaluation servers or toolkits.
As shown in~\cref{tab:main_result}, when compared to VOS specialist models, \modelname achieves competitive performance on VOS benchmarks, even without seeing any training videos. Furthermore, in comparison to segmentation generalist models, \modelname surpasses Painter~\cite{painter}, SEEM~\cite{seem}, and PerSAM~\cite{persam} by a significant margin, and competes favorably with recent generalist models~\cite{seggpt,matcher}. Additionally, the VOS task pipeline in \modelname is relatively simple. We do not use any test time augmentation tricks (TTA) used in~\cite{xmem}. Moreover, it does not involve dense feature interaction at the patch level, as in SegGPT, and does not require the use of a pre-trained SAM for segmentation.

\vspace{0.05in}
\noindent\textbf{Generic semantic segmentation.}
We also evaluate \modelname on generic semantic segmentation benchmarks,  which need to segment dataset-dependent pre-defined categories within each image.
We adapt generic semantic segmentation to our in-context learning framework by gathering in-context examples from the training set and then performing segmentation in an in-context manner.
We use two widely-used semantic segmentation datasets, COCO~\cite{coco} and ADE20k~\cite{ade20k}, for evaluation.
As illustrated in~\cref{tab:main_result}, compared to specialist and generalist models for semantic segmentation, \modelname demonstrates strong performance under the settings that the classes are known in advance (marked with \#). As shown in~\cref{tab:main_result}, our best model achieves 74.0 and 59.0 mIoU on COCO and ADE20k, respectively, surpassing the previous specialist/generalist models.

\vspace{0.05in}
\noindent\textbf{Open-vocabulary semantic segmentation.}
Similar to generic semantic segmentation, we can also extend \modelname to open-vocabulary semantic segmentation. 
Furthermore, we could also utilize StableDiffusion~\cite{sd} to synthesize images with coarse masks for those categories lacking in-context examples like~\cite{nguyen2023dataset}.
Compared to X-decoder~\cite{x-decoder}, our method obtains competitive and even better results on PC-459 and ADE-847.

\begin{table}[!h]
      \centering
      \tablestyle{1.5pt}{1.2}
      \scriptsize
    \caption{ \textbf{ Ablation on foundation models as backbone.}
    HRes denotes high-resolution pre-training; MRes denotes multi-resolution pre-training or intrinsic support within the architecture, \eg CNN and UNet; Task$^p$ denotes the pre-training task.
    \colorrect{Gray} means that we only obtain a trivial performance.}
    \vspace{-3mm}
      \resizebox{\linewidth}{!}{
      \begin{tabular}{l|lcc|lr| ccc|c}
      Encoder & Arch & HRes & MRes & Task$^p$ & Data$^p$ & COCO-20$^i$ & FSS-1000 & DAVIS-17 & AVG\\
      \shline
      DINOv2-B~\cite{dinov2} &ViT&\Checkmark&\Checkmark& image/patch self-distillation & 142M & 74.6 & 87.4 & 68.4 &76.8 \\
      DINOv2-L~\cite{dinov2} &ViT&\Checkmark&\Checkmark& image/patch self-distillation & 142M & 75.4 & 86.9 & 68.9 & 77.1\\
      DINOv2-G~\cite{dinov2} &ViT&\Checkmark&\Checkmark& image/patch self-distillation & 142M & 75.7 & 87.5 & 70.1 & 77.8 \\
      \hline
      DINOv1-B~\cite{dino} &ViT& &\Checkmark& image self-distillation &1M & 53.5 & 76.9 & 54.2 & 61.5\\
      SAM-B~\cite{sam} &ViT&\Checkmark& & interactive segmentation & 11M & 59.8 & 76.9 & 60.1 & 65.6\\
      SD-2.1~\cite{sd} &UNet&\Checkmark&\Checkmark& text-to-image generation & 2B & 66.6 & 84.2 & 52.4 & 67.7\\
      OpenCLIP-ConvNext-B~\cite{open_clip} &CNN& &\Checkmark& image-text alignment & 400M & 65.1 &77.0 & 50.4 & 64.2\\
      \rowcolor{Gray}
      CLIP-ViT-B~\cite{clip} &ViT& & & image-text alignment & 400M & 15.3 & 29.4 & 1.3 & 15.3\\
      \rowcolor{Gray}
      MAE-B~\cite{mae} &ViT& & & mask image modeling & 1M & 12.0 & 23.7 & 2.3 & 12.6 \\
      \hline
      ConvNext-B~\cite{convnext}&CNN& &\Checkmark& image classification & 1M & 62.0 & 72.8 & 37.6 & 57.5\\
      \end{tabular}}
    \label{tab:abl_backbone}
    \vspace{-5mm}
\end{table}
\begin{figure}[!h]

    \centering
    \includegraphics[width=0.98\linewidth]{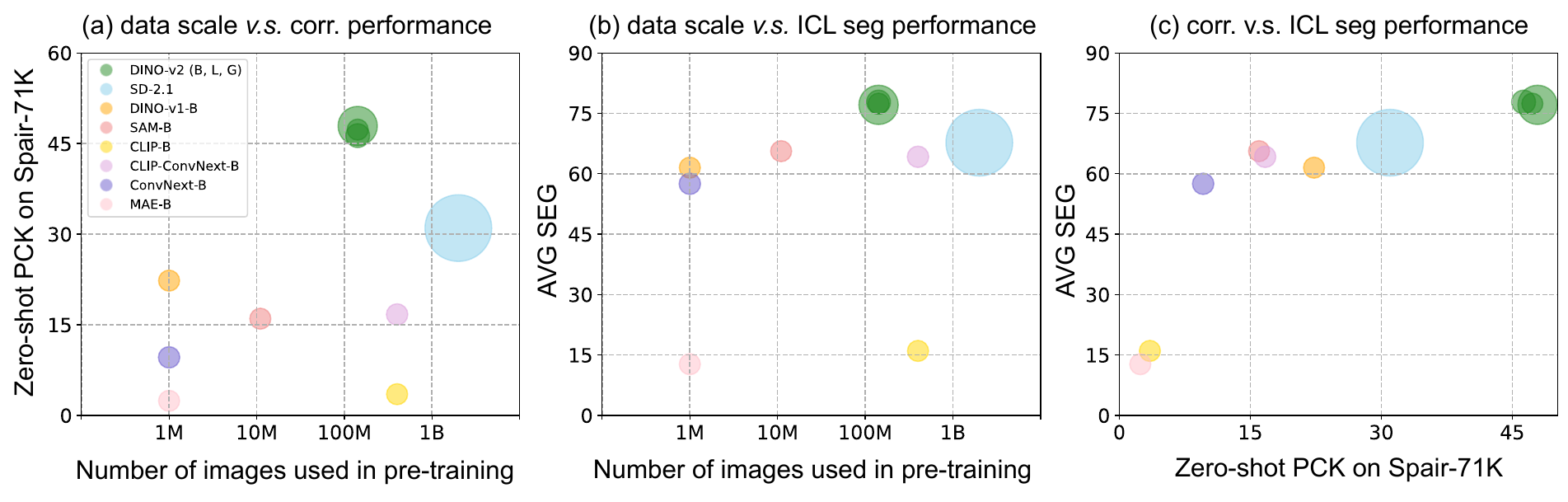}
    \vspace{-4mm}
    \caption{\textbf{Performance on zero-shot semantic correspondence and in-context segmentation.} We use the dense feature similarity for correspondence estimation. The diameter of each bubble represents the number of parameters of each model.}
    \label{fig:corr_seg}
    \vspace{-8mm}
\end{figure}
\vspace{0.05in}

\subsection{Rethinking Vision Foundation Model Pre-training}
Existing works improve VFMs by scaling up data and model sizes, modifying the model architectures, and utilizing different pre-text tasks. It is unknown how much those factors affect the performance when the encoder is frozen. We delve deeper into the potential of various \textbf{frozen} pre-trained VFMs~\cite{dino, dinov2, mae, sam, sd, clip, open_clip, convnext}.
For each VFM, we freeze it to extract image features and dense correspondences then map them into the hidden space for mask decoding, equally. 
Additionally, we assess the \textbf{zero-shot} semantic correspondence score to underscore the emerging downstream capabilities from different perspectives directly. More specifically, we evaluate a percent of correct keypoints score (PCK) on SPair-71k~\cite{spair} following to~\cite{cats++} with cosine similarity between the dense feature maps of the two images for semantic correspondence.
We observe that (1) high/multi-resolution is essential; (2) pre-text tasks are important; (3) model and data scales do not necessarily help for ICL segmentation; (4) zero-shot correspondence can imply the ICL segmentation capacity.

\vspace{0.05in}
\noindent\textbf{High/multi-resolution is essential.}
It is evident in~\cref{tab:abl_backbone} that the models (CLIP-ViT-B and MAE-B), lacking support for high-resolution or multi-resolution only exhibit trivial performance within our frozen-backbone framework. 
Despite the remarkable success of vision transformers, they encounter limitations in directly adapting to various input resolutions without fine-tuning due to fixed-length visual tokens and position embeddings.
Unlike vision transformers, CNNs, with their stacked convolution design, demonstrate seamless adaptation of parameters to images of different resolutions, even only pre-trained on low-resolution images.
To alleviate this limitation, DINO-v1/v2 enhance vanilla ViTs with multi-resolution inputs and position embedding interpolation during pre-training. Consequently, they achieve superior segmentation performance.
Overall, the observations indicate that the high/multi-resolution capabilities of the pre-trained VFMs play a key role in in-context segmentation.

\vspace{0.05in}
\noindent\textbf{Pre-text tasks are important.}
Once the conditions for supporting high-resolution inputs are met, pre-text tasks employed during pre-training and the scale of data also impact the in-context segmentation capacities. Illustrated in~\cref{fig:corr_seg}, with the same model size and pre-training data, DINOv1 outperforms those classification-pre-trained models in both zero-shot semantic correspondence and in-context segmentation tasks by a large margin. This emphasizes the substantial influence of the pre-text task on downstream capabilities.

\vspace{0.05in}
\noindent\textbf{Model and data scales do not directly aid ICL segmentation.}
Equipping with larger-scale data and advanced self-distillation techniques, DINOv2 outperforms DINOv1 by a considerable margin. 
However, as shown in~\cref{tab:abl_backbone}, scaling up the model size from base (B) to giant (G), is not a clear performance boost.
Furthermore, compared to those pre-trained on than 10 times data, \eg OpenCLIP-ConvNext-B and SD-2.1, DINOv2 still exhibits significant advantages.
Those demonstrate that model and data scales do not directly help.

\vspace{0.05in}
\noindent\textbf{Zero-shot correspondence implies the segmentation capacity.}
We further study the relationship between correspondence and segmentation performance with different backbones. 
As shown in~\cref{fig:corr_seg}, the performance on segmentation is proportional to correspondence, demonstrating a strong consistency between these two tasks. It further confirms our motivation to leverage the emergent correspondence for in-context segmentation.
This insight further inspires us, suggesting that pre-training emphasizing inter/intra-image correspondence may lead to better in-context segmentation potentials, \eg the image/patch-level discriminative self-distillation~\cite{dinov2}.

\subsection{Ablation Study}
In the ablation study, we report the performance on COCO-20$^i$~(in-domain one-shot segmentation), FSS-1000~(out-of-domain one-shot segmentation) and DAVIS-17~(zero-shot video object segmentation) to investigate the in-domain convergence and out-of-domain generalization capability.

\vspace{0.05in}
\hspace{-6mm}
\begin{minipage}{\linewidth}
\begin{minipage}[t]{0.48\textwidth}
\makeatletter\def\@captype{table}
\caption{ \textbf{Component ablation on in-context instructions.} 
\colorrect{lightblue} is the default setting for in-context instructions.}
\vspace{-3mm}
\tablestyle{1.5pt}{1.2}
\scriptsize
\resizebox{\linewidth}{!}{
  \begin{tabular}{ccc| c |c|c}
  geometric & visual & meta & COCO-20$^i$ & FSS-1000 & DAVIS-17\\
  \shline
    \Checkmark & & & 68.8 & 87.3 & 65.8 \\
  & \Checkmark & & 71.0 & 85.1 & 67.6\\
  & & \Checkmark  &  73.7 & 63.7 & 21.6\\
  \hline
  &&& 48.9 & 58.1 & 21.1 \\
  \rowcolor{lightblue}
  \Checkmark & \Checkmark & \Checkmark & \textbf{74.6} & \textbf{87.4} & \textbf{68.4} \\
  \end{tabular}
\label{tab:abl_comp}}
\end{minipage}
\hfill
\begin{minipage}[t]{0.48\textwidth}
\makeatletter\def\@captype{table}
    \caption{ \textbf{Ablations the training strategies.} 
    \colorrect{lightblue} is the default training scheme in our main experiments.}
    \vspace{-3mm}
      \tablestyle{1.5pt}{1.2}
      \scriptsize
    \resizebox{\linewidth}{!}{
      \begin{tabular}{cc| c|c|c}
      reversion & negative & COCO-20$^i$ & FSS-1000 & DAVIS-17\\
      \shline
      &  &  70.0 & 86.1 & 66.3\\
       & \Checkmark  & 72.9 & 86.9 & 65.0\\
      \Checkmark & & 73.1 & 86.5 & 64.3\\
      \hline
      \rowcolor{lightblue}
      \Checkmark & \Checkmark  & \textbf{74.6} & \textbf{87.4} & \textbf{68.4}\\
      \end{tabular}
    \label{tab:abl_tricks}}
\end{minipage}
\vspace{3mm}
\end{minipage}

\vspace{0.05in}
\noindent\textbf{Ablations on in-context instructions.}
We study the importance of each component of in-context instructions (\ie geometric, visual, and meta instructions). We conduct ablations on different combinations of components.
As shown in~\cref{tab:abl_comp}, we find that the geometric and visual instructions tend to help the performance for out-of-domain segmentation and meta instructions (class name and task description in the ablation) benefit in-domain performance: when each component is used individually, the geometric and visual instructions obtain the best results on FSS-1000 and DAVIS-17, respectively. Meanwhile, meta instructions achieve the best performance on COCO-20$^i$, but with poor results on the other two datasets. 
Encouragingly, our method obtains the best performance among three tasks when using all three prompts, further demonstrating that our model can effectively transfer knowledge from in-context samples with the proposed in-context instructions.

\vspace{0.05in}
\noindent\textbf{Ablations on training strategies.} We investigate how the proposed training strategies affect performance. As shown in~\cref{tab:abl_tricks}, since the two strategies, \ie context reversion and negative entity augmentation, act as a form of ``in-context augmentation'', they bring performance gains on COCO-20$^i$ and FSS-1000$^i$ with a slight drop on DAVIS-17. Furthermore, after combining them, our model achieves a consistent gain across all datasets, highlighting its efficacy. 

\begin{table*}[h]
    \vspace{-5mm}
      \centering
      \tablestyle{1.5pt}{1.2}
      \scriptsize
    \caption{\textbf{Ablations on training data.} 
    \colorrect{lightblue} is the default data combination.}
    \vspace{-3mm}
      \begin{tabular}{cc |cc | ccc | c }
      \multicolumn{2}{c}{\textit{semantic segmentation}} & \multicolumn{2}{c}{\textit{instance segmentation}} & \multirow{2}{*}{COCO-20$^i$} & \multirow{2}{*}{FSS-1000} & \multirow{2}{*}{DAVIS-17}  & \multirow{2}{*}{AVG}\\
      COCO$_\mathrm{sem}$& ADE20k  & COCO$_\mathrm{inst}$ & LVIS  &&&&
      \\
      \shline
      \Checkmark &  & &  & \textbf{76.3} & 80.9 & 45.1 & 67.4\\
      \Checkmark & \Checkmark   & & &75.6 & 82.3 & 40.1 & 66.0 \\
      \hline
      \Checkmark &  & \Checkmark &  &  75.7 & 83.5 & \textbf{70.1} & 76.4\\ 
      \rowcolor{lightblue}
      \Checkmark & \Checkmark  & \Checkmark & \Checkmark   & 74.6 & \textbf{87.4} &  68.4 & \textbf{76.8}\\
      \end{tabular}
    \label{tab:abl_dataset}
    \vspace{-7mm}
\end{table*}

\vspace{0.05in}
\noindent\textbf{Dataset Combination.}
We further investigate the effectiveness of each dataset under our joint in-context training framework. We group the training data into two types: semantic segmentation (COCO$_\mathrm{sem}$, ADE20k) and instance segmentation (COCO$_\mathrm{inst}$, LVIS). As shown in~\cref{tab:abl_dataset}, it is clear to see that when trained on COCO$_\mathrm{sem}$ only, it achieves the best performance on COCO-20$^i$ but is weak on the other. 
Based on this, we can enrich the training data from two directions: (1) use extra semantic segmentation data and (2) introduce instance segmentation tasks into training. For the former direction, the performance on FSS-1000 increases after introducing ADE20k. For the latter, we still use COCO as the training set but introduce its instance-level annotation. It can be seen a clear performance boost on DAVIS-17 $\mathcal{J} \& \mathcal{F}$ score (45.1$\to$70.1), since video object segmentation requires instance-level understanding. Finally, we further enrich our training data with LVIS~\cite{lvis}, achieving the best overall performance.

\section{Conclusion}
We introduced \modelname, an end-to-end in-context segmentation framework that leverages the emergent correspondence of a single frozen vision foundation model. By training on standard segmentation datasets, \modelname achieved \sota performance on one-shot segmentation benchmarks. Impressively, \modelname demonstrated competitive performance on novel tasks, providing a cost-effective training approach for universal segmentation. In this work, our primary focus is on utilizing one in-context example per entity. In the future, we plan to explore utilizing multiple in-context examples to enhance contextual information. Additionally, we aim to investigate our model's potential in instance-level segmentation, such as open-world instance segmentation.
We do not anticipate any undesirable ethical or social impacts.

{
\small
\noindent \textbf{Acknowledgement} This project was supported by NSFC under Grant No. 62032006 and No. 62102092. We appreciate the valuable feedback from  Xinlong Wang.
}

\newpage
\bibliographystyle{splncs04}
\bibliography{refs}

\newpage
\appendix

\section{Additional Implementation Details}
In this section, we provide additional details of the mask decoding and training pipeline.

\vspace{0.05in}
\noindent\textbf{Mask decoding.} Inspired by recent query-based segmentation methods~\cite{detr,maskformer,mask2former,sam}, we utilize a lightweight query-based mask decoder to effectively map the in-context enhanced image features and object query to an output mask. We employ a learnable object query that will be used for the decoder’s output.
The workflow of \modelname is concisely summarized in Pytorch-style pseudocode, as presented in~\cref{alg:mask_decoding}. Initially, the image feature and in-context instructions are projected into the same feature space for mask decoding. Subsequently, the projected in-context features are leveraged to enhance the image feature and object feature. 
The mask decoder then executes a multi-layered decoding process, structured as a four-step procedure within each layer, including (1) self-attention for the concatenated object query; (2) cross-attention from the image feature to the object query; (3) cross-attention from the query back to the image; and (4) calculating the mask. The mask calculation is performed using the image feature and the first element of the concatenated object feature, which corresponds to the position of the initial object query.

\vspace{0.05in}
\noindent\textbf{Training pipeline.}
In our main experiments, we adopt a mixed training scheme using both semantic segmentation datasets (COCO and ADE20k) and instance segmentation datasets (COCO and LVIS), as presented in  Algorithm~\ref{alg:training_pipeline}. We do not focus intensively on adjusting the dataset ratios, instead opting for uniform dataset-level sampling. For the segmentation datasets, we employ large-scale jittering (ranging from 0.1 to 2.0) for both the target image and in-context examples. These in-context examples are constructed based on the semantic class label of the target image, sampling one class per image during training. In the case of instance segmentation datasets, in-context examples are generated by applying two separate data augmentations to the same image. The instances from these differently augmented views then serve as mutual in-context examples. Our standard data augmentation techniques for this task include random resizing cropping (ranging from 0.3 to 1.0), random color jittering (with a 0.2 probability), and random horizontal flipping (with a 0.1 probability).

\section{Additional Visualization}
In this section, we provide more visualizations of the middle output and the predictions of \modelname.

\begin{figure}[!ht]
    \centering
    \includegraphics[width=\linewidth]{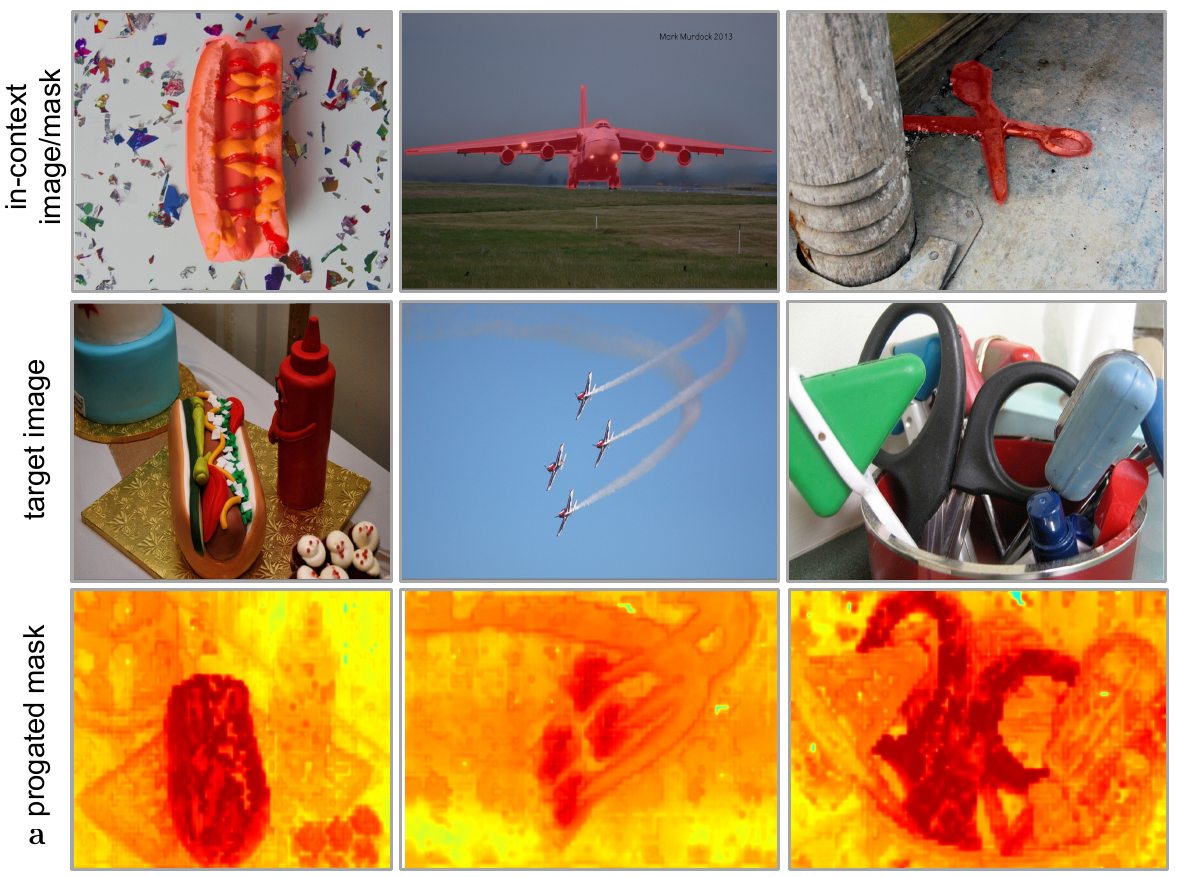}
    \vspace{-0.1in}
    \caption{\textbf{Visualization of propagated masks.} We propagate labels from the in-context examples to the targets to obtain propagated masks by exploring the dense correspondences. We employ DINO-v2-large~\cite{dinov2} for the visualization.}
    \label{fig:vis_heat}
\end{figure}

\vspace{0.05in}
\noindent\textbf{Propagated mask.}
As outlined in Section 3.2, the propagated mask $\mathbf{a}$ is derived from a weighted mean of dense correspondences according to the ground-truth mask of in-context samples. To facilitate visualization, we first apply the sigmoid function to map $\mathbf{a}$ into $(0, 1)$. Subsequently, this range is transformed into RGB space using the JET color map. As depicted in~\cref{fig:vis_heat}, this process demonstrates that the propagated masks predominantly concentrate on the objects referenced in the in-context examples, providing strong guidance for the subsequent mask decoding process. 
This observation further demonstrates the emerging potential of pre-trained vision foundation models in the realm of segmentation tasks.

\begin{figure*}[!ht]
  \begin{minipage}{\linewidth}
    \myalgorithm
  \end{minipage}
\vspace{-5pt}
\end{figure*}

\begin{figure*}[!ht]
  \begin{minipage}{\linewidth}
    \myalgorithmtwo
  \end{minipage}
\vspace{-5pt}
\end{figure*}

\begin{figure*}[!h]
    \centering
    \includegraphics[width=\linewidth]{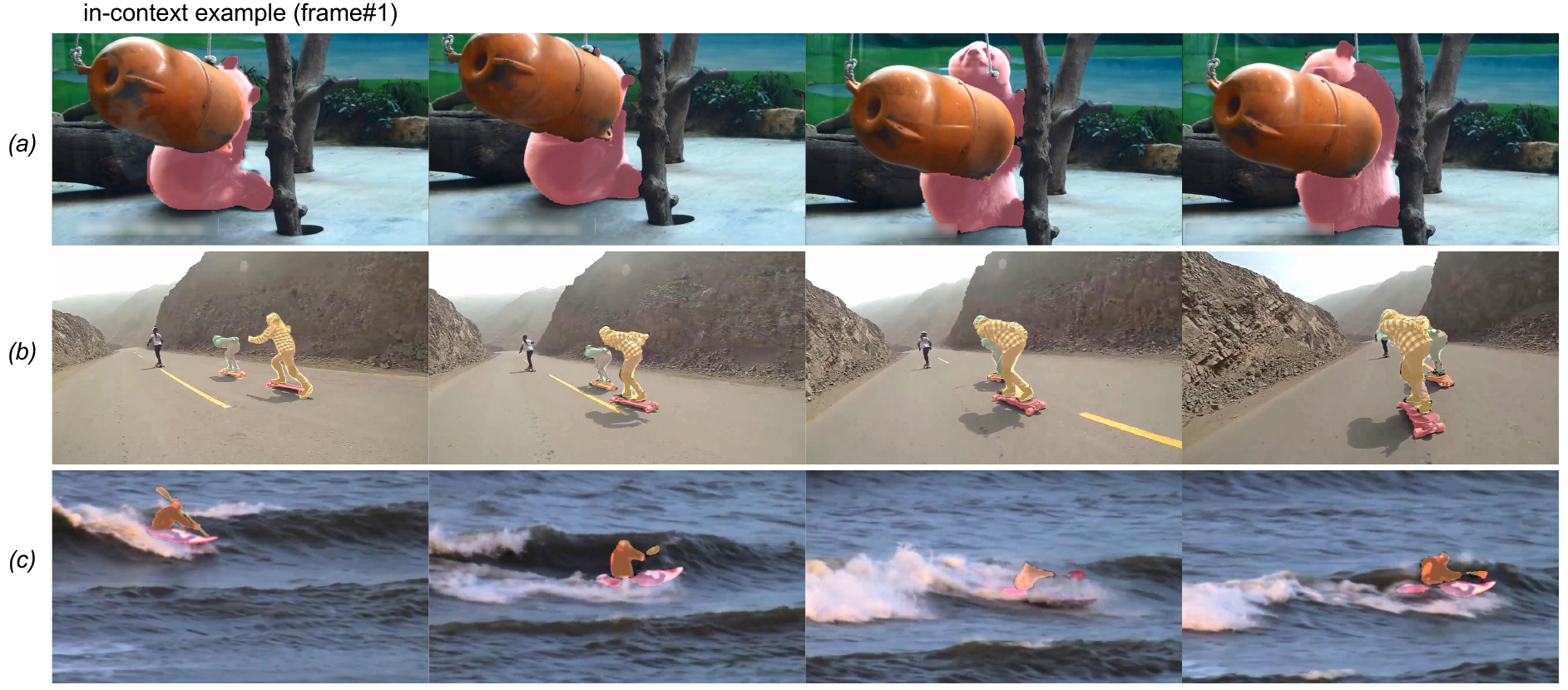}
    \vspace{-0.1in}
    \caption{\textbf{Qualitative results on VOS.} \modelname perform well on challenging scenarios in video object segmentation, including (a) occlusions, (b) interwoven objects, and (c) small objects.}
    \label{fig:vis_vos}
\end{figure*}

\vspace{0.05in}
\noindent\textbf{More qualitative results on VOS.}
We further provide more qualitative results on video object segmentation tasks (VOS). Note that \modelname is never trained on video datasets and just treats VOS as in-context segmentation using the first frame as in-context examples. As shown in~\cref{fig:vis_vos}, \modelname well handles challenging scenarios, including (a) occlusions, (b) interwoven objects, and (c) small objects.


\end{document}